\documentclass{llncs}
\usepackage{algorithm}
\usepackage{algorithmic}
\usepackage{graphicx}
\usepackage{hyperref}
\begin{document}
\title{Learning From How Humans Correct}
%
%
\author{
Tong Guo
}
%

%
\institute{779222056@qq.com}
\maketitle              
\begin{abstract}

In industry NLP application, our manually labeled data has a certain number of noisy data. We present a simple method to find the noisy data and relabel them manually, meanwhile we collect the correction information. Then we present novel method to incorporate the human correction information into deep learning model. Human know how to correct noisy data. So the correction information can be inject into deep learning model. We do the experiment on our own text classification dataset, which is manually labeled, because we need to relabel the noisy data in our dataset for our industry application. The experiment result shows that our learn-on-correction method improve the classification accuracy from 91.7\% to 92.5\% in test dataset. The 91.7\% accuracy is trained on the corrected dataset, which improve the baseline from 83.3\% to 91.7\% in test dataset. The accuracy under human evaluation achieves more than 97\%.

\end{abstract}
\section{Introduction}

In recent years, deep learning \cite{ref_proc2} and BERT-based \cite{ref_proc1} model have shown significant improvement on almost all the NLP tasks. However, past methods did not inject human correction information into the deep learning model. Human interact with the environment and learn from the feedback from environment to correct the their own error or mistake. Our method try to solve the problem that let deep learning model imitate how human correct.

In order to solve the problem we present a learning framework. The framework mainly works for our industry application dataset, because the problem starts from our industry application. In order to solve the text classification problem in our industry application, we first label a dataset. Then we can find the noisy data in the dataset and relabel them. The relabeling step collects the human correction information.

To the best of our knowledge, this is the first study exploring the improvement of injecting human correction information into deep model for natural language understanding. Our key contribution are 3 folds:

1. Based on our dataset, we first present the simple step to find the noisy data and relabel the noisy data.

2. We present the method to inject the human correction information into BERT for text classification.  

3. Our learning framework can apply to a broad set of deep learning industry applications whose dataset is manually labeled.
 
\begin{figure}
\centering
\includegraphics[width=\textwidth]{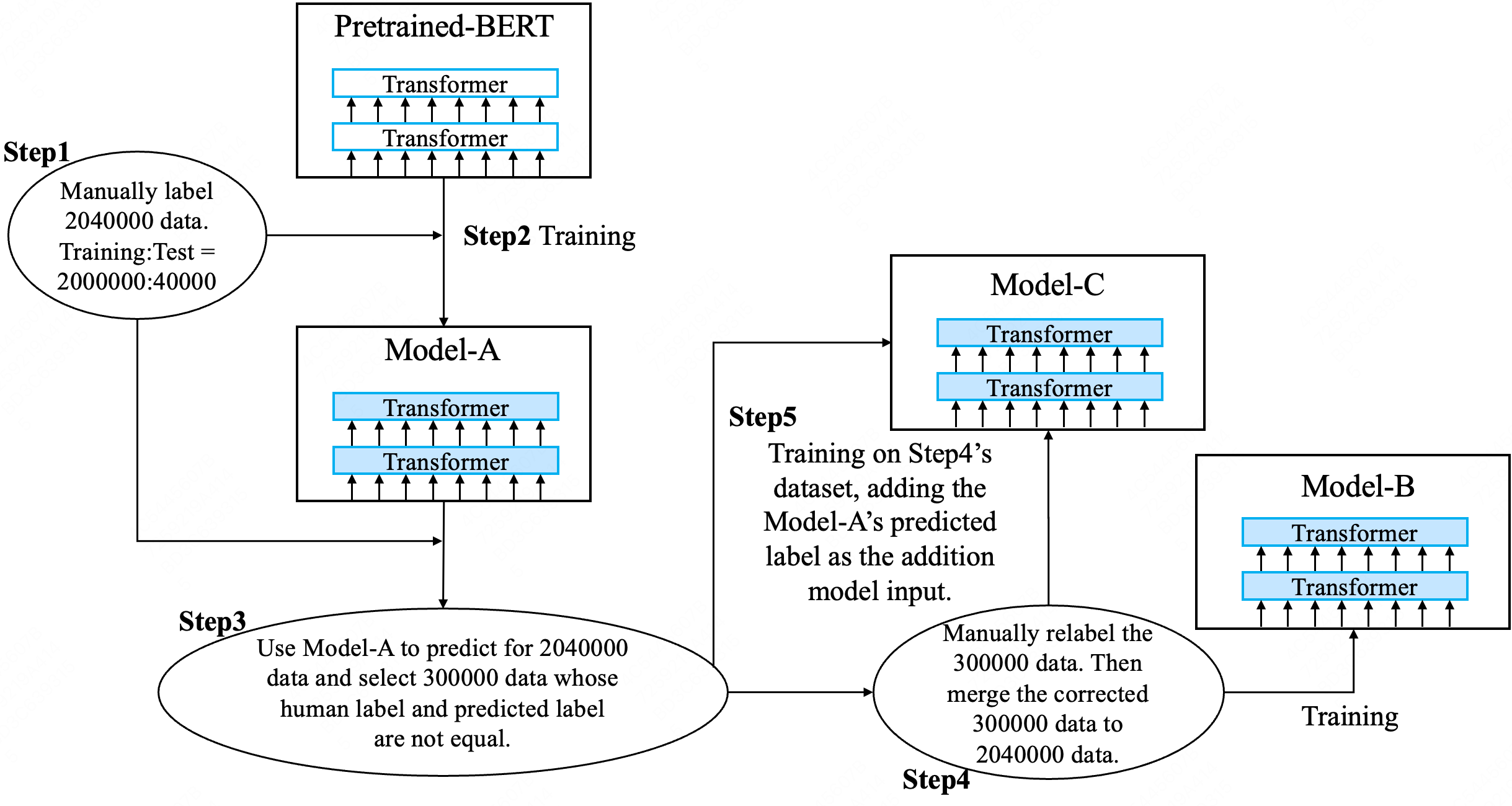}
\caption{Our learning framework} \label{fig1}
\end{figure}

\section{Relate Works}

BERT \cite{ref_proc1} is constructed by the multi-layer transformer encoder \cite{ref_proc10}, which produces contextual token representations that have been pre-trained from unlabeled text and fine-tuned for the supervised downstream tasks. BERT achieved state-of-the-art results on many sentence-level tasks from the GLUE benchmark \cite{ref_proc3}. There are two steps in BERT's framework: pre-training and fine-tuning. During pre-training, the model is trained on unlabeled data by using masked language model task and next sentence prediction task. Apart from output layers, the same architectures are used in both pre-training and fine-tuning. The same pre-trained model parameters are used to initialize models for different down-stream tasks.

Our method is different to semi-supervised learning. Semi-supervised learning solve the problem that making best use of a large amount of unlabeled data. These works include UDA \cite{ref_proc6}, Mixmatch \cite{ref_proc7}, Fixmatch \cite{ref_proc8}, Remixmatch \cite{ref_proc9}. These works do not have the human correction information. Our work has a clear goal that is to learn how humans correct their mistake. \cite{ref_proc11} studies the PN learning problem of binary classification. But our idea can apply to a broad set of deep learning industry applications.

The work\cite{ref12} of OpenAI also use model predictions for human as references to label. The work\cite{ref12} use the new human-labeled data from model predictions to train the reward model, which is reinforcement learning from human feedback(RLHF). Our work does not have the reward model of RLHF, the differnce between RLHF and our relabel method is that RLHF focuses on using the reward-model/reward-dataset to guide policy-model/policy-dataset, and our relabel method focus on correcting the policy dataset. Also, the new human-labeled data for reward-model does not conflict to our method, because the new human-feedback reward dataset can be simply merged to the policy dataset, and then to relabel/correct by our method.

Also, correcting the \cite{ref12}'s policy by reward model is same to correcting all the related data/labels in policy training dataset. And the detail pattern-based/substring-based human-correct method is illustrated at \cite{ref13}. \cite{ref13} uses the human-feedback data to correct the policy training dataset. In summary, our method focuses on getting a high quality policy dataset.

Confident learning (CL)\cite{ref14} developed a novel approach to estimate the joint distribution of label noise and explicated theoretical and experimental insights into the benefits of doing so. CL demonstrated accurate uncertainty quantification in high noise and and sparsity regimes, across multiple datasets, data modalities, and model architectures. CL is a data-centric\cite{ref15} approach which focuses instead on label quality by characterizing and identifying label errors in datasets, based on the principles of pruning noisy data, counting with probabilistic thresholds to estimate noise, and ranking examples to train with confidence. But our method is more practical and simple.

Noisy label detection is closely related to the relabel step of our framework. Hendrycks and Gimpel \cite{ref16} use the maximum softmax probability as a baseline to detect misclassified and out-of-distribution examples: a low top-class score means the model is uncertain about that example. Kuan and Mueller \cite{ref17} propose a model-agnostic label quality score to find real-world label errors. Yu et al. \cite{ref18} study noisy label detection when a small set of clean data is available. These methods focus on how to find noisy labels. Our method uses a simple disagreement rule: if the model prediction and the human label are not equal, we treat the example as noisy and relabel it by hand. Another uncertainty signal in this line of work is that the top-1 predicted score and the top-2 predicted score are very close, which means the model can hardly classify the example and the label is more likely to be noisy. We leave the combination of our relabel method and these detection methods, and the search for the most efficient one, to future work.

\section{Our Method}

In this section, we describe our method in detail. Our learning framework is shown in Fig 1. The framework includes 5 steps:

Step 1, in order to solve the industry text classification problem. We label 2,040,000 data and split them into 2,000,000 training data and 40,000 test data. The 2,040,000 data are sampled from our application database whose data size is 500 million. 

Step 2, we train / fine-tune the BERT model on the 2,000,000 training data. We named the result model of this step Model-A. Note that Model-A should not overfit the training dataset.

Step 3, we use Model-A to predict for all the 2,040,000 data.  Then we find 300,000 data whose predicted label and human label are not equal. We consider it is the noisy data. In detail, there are 294,120 noisy data in the training dataset and 5,880 noisy data in the test dataset.

Step 4, we manually relabel 300,000 data and merge back to the 2,040,000 data. Then we get the merged 2,040,000 data. During the relabeling, the last label by human and the model-A's predicted label are listed as the references for people. But labeling people also make their own decision.    

Step 5, we add the Model-A's predicted one-hot label as the addition input for training / fine-tuning a new model. The detail encoding method for the predicted label is described in the next section. We named the result model of this step Model-C. The Model-A's predicted one-hot labels represent the before-corrected information. The training ground truth for Model-C contains the 294,120 corrected human label, which represent the corrected information. So Model-C is learning how to correct and learning the text classification task in the same time.

\begin{figure}
\centering
\includegraphics[width=\textwidth]{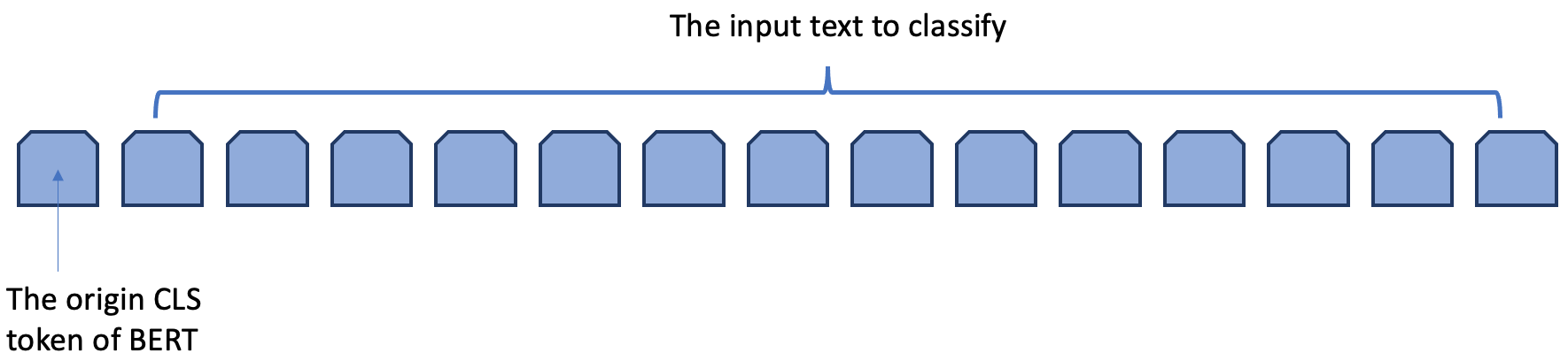}
\caption{The encoding detail for Model-A, which is corresponding to Fig 1.} \label{fig2}
\end{figure}

\begin{figure}
\centering
\includegraphics[width=\textwidth]{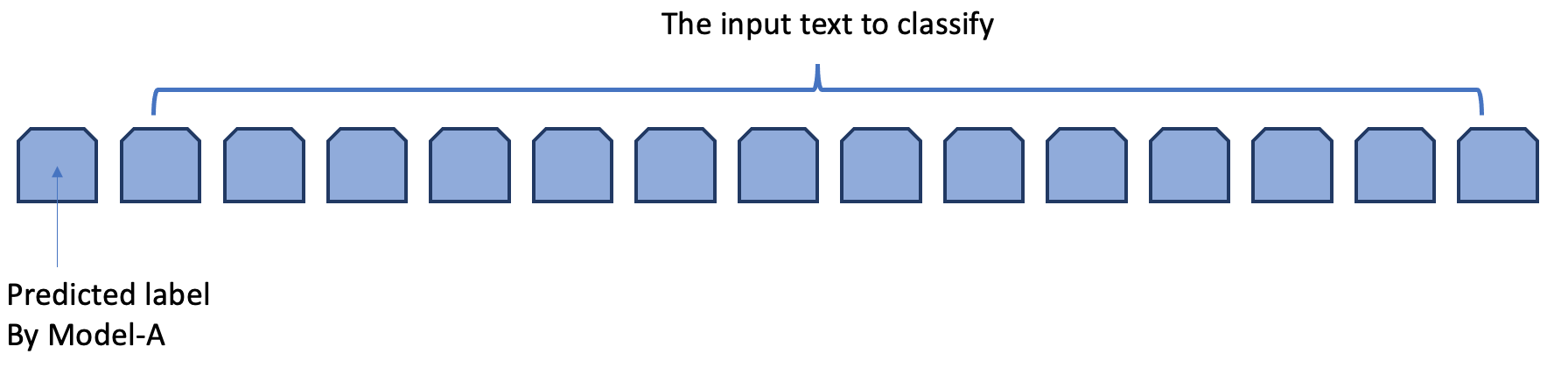}
\caption{The encoding detail for Model-C, which is corresponding to Fig 1.} \label{fig3}
\end{figure}

\section{The Model}

We use BERT as our model. The training steps in our method belongs to the fine-tuning step in BERT. We follow the BERT convention to encode the input text. The encoding detail for Model-A is shown in Fig 2. The encoding detail for Model-C is shown in Fig 3.

\section{Experiments}

In this section we describe detail of experiment parameters and show the experiment result. The detail result is shown in Table 2. The data size in our experiment is shown in Table 1.

In fine-tuning, we use Adam \cite{ref_proc4} with learning rate of 1e-5 and use a dropout \cite{ref_proc5} probability of 0.1 on all layers. We use BERT-Base (12 layer, 768 hidden size) as our pre-trained model. 

\begin{table}
\caption{The data size for text classification}\label{tab1}
\centering
\begin{tabular}{|l|l|l|}
\hline
Data Size & Description \\
\hline
500,000,000 &  All the data in our application database. \\ 
\hline
2,040,000 & The data we label in step 1 of Fig 1. \\ 
\hline
2,000,000 & The the training dataset we split from the 2,040,000 data. \\ 
\hline
300,000 & All the noisy data we select from the 2,040,000 data to relabel. \\ 
\hline
294,120 & The noisy data in the training dataset. \\ 
\hline
40,000 & The test dataset we split from the 2,040,000 data. \\
\hline
5,880 & The noisy data in the testing dataset. \\ 
\hline
\end{tabular}
\end{table}

\begin{table}
\caption{The experiment result of text classification. The test dataset is the 40,000 data. The accuracy reported by human is 5000 data that sampled from the 500 million data. Model-A, Model-B and Model-C are corresponding to Fig 1.}\label{tab1}
\centering
\begin{tabular}{|l|l|l|}
\hline
Model In Fig 1 & Test Dataset Accuracy & Human Evaluate Accuracy \\
\hline
Model-A & 83.3\% & 88.0\% \\ 
\hline
Model-B & 91.7\% & 97.2\% \\ 
\hline
Model-C & 92.5\% & 97.7\% \\ 
\hline
\end{tabular}
\end{table}

\section{Discussion}
In step 4 of Fig 1, the manually relabeling can correct the noisy 300,000 data. Because the selected 300,000 data is unfitting to the Model-A, the relabeling's 'worst' result is the origin last human's label.

The step 5 in Fig 1 is the core contribution of our work. In step 5, the predicted label of Model-A contains the before-human-corrected information. The ground truth for Model-C contains the after-human-corrected information. So the model is learning the human correction.

We could use the before-corrected human label (i.e., the ground truth in step 1 of Fig 1) as the input for Model-C. But this way can not apply to real industry inference. Because we can not get the before-corrected human label as the input of Model-C in real industry application.  In real industry application, we use the Model-A to predict one-hot label as input for Model-C. 

Human evaluation accuracy is higher than the test dataset accuracy, because we randomly sampled 5000 data from the 500 million data. The sampled 5000 data represents the great majority of the 500 million data.

\textbf{Why relabel method work?} Because deep learning is statistic-based. Take classification as example. (In a broad sense, all the machine learning tasks can be viewed as classification.) If there are three \textbf{very similar} data (data-1/data-2/data-3) in total, which labels are class-A/class-A/class-B. Then the trained model will probably predict class-A for data-3. We assume that data-3 is wrong labeled to class-B by human , because more people label its similar data to class-A. If we do not correct data-3, the model prediction for new data that is the most similar to data-3 will be class-B, which is wrong. The new data is more similar to data-3 than data-1/data-2.

The cost of annotation time is crucial for deep learning tasks based on manually labeled data. If there is not enough labeling manpower, we must find ways to reduce the amount of data that needs to be labeled. If we can reduce the amount of data to be labeled to a level that a single programmer can handle, then we do not need an additional labeling team.

\section{Conclusion}

Humans interact with the environment and learn from the feedback to correct their own errors. Based on this correction idea, we design a learning framework to inject the information of human correction into the deep model. The experiment result shows that our idea works. For further improvement, we will relabel the data whose top-1 and top-2 predicted scores are very close, and combine our relabel method with the noisy data detection methods discussed above. Our idea can apply to a broad set of deep learning industry applications.

\end{document}